\documentclass{article}

\usepackage[preprint]{neurips_2024}

\usepackage[utf8]{inputenc} 
\usepackage[T1]{fontenc}    
\usepackage{hyperref}       
\usepackage{url}            
\usepackage{booktabs}       
\usepackage{amsfonts}       
\usepackage{nicefrac}       
\usepackage{microtype}      
\usepackage{xcolor,colortbl}
\definecolor{MyPink}{RGB}{255, 221, 219}
\definecolor{MyBlue}{RGB}{0, 0, 255}
\usepackage{amsmath}
\usepackage{amssymb}
\usepackage{algorithm}
\usepackage{algpseudocode}
\usepackage[edges]{forest}
\usepackage{stfloats}
\usepackage{multirow}
\usepackage{wrapfig}
\setcitestyle{square,comma,numbers}
\usepackage{array}          
\newcommand\ChangeRT[1]{\noalign{\hrule height #1}}

\title{ELSA: Exploiting Layer-wise N:M Sparsity for \\Vision Transformer Acceleration}

\author{
Ning-Chi Huang$^1$,~
Chi-Chih Chang$^1$,~
Wei-Cheng Lin$^1$, \\
\textbf{Endri Taka$^2$,~
Diana Marculescu$^2$,~
and Kai-Chiang Wu$^1$}
\and
$^1$National Yang Ming Chiao Tung University, $^2$University of Texas at Austin\\
{\tt\small nchuang@cs.nycu.edu.tw, brian1009.en08@nycu.edu.tw, weicheng.lin.cs11@nycu.edu.tw, }\\
{\tt\small endri.taka@utexas.edu, dianam@utexas.edu, kcw@cs.nctu.edu.tw}
}

\begin{document}
\maketitle

\begin{abstract}
$N{:}M$ sparsity is an emerging model compression method supported by more and more accelerators to speed up sparse matrix multiplication in deep neural networks. Most existing $N{:}M$ sparsity methods compress neural networks with a uniform setting for all layers in a network or heuristically determine the layer-wise configuration by considering the number of parameters in each layer. However, very few methods have been designed for obtaining a layer-wise customized $N{:}M$ sparse configuration for vision transformers (ViTs), which usually consist of transformer blocks involving the same number of parameters.
In this work, to address the challenge of selecting suitable sparse configuration for ViTs on $N{:}M$ sparsity-supporting accelerators, we propose ELSA, Exploiting Layer-wise $N{:}M$ Sparsity for ViTs. Considering not only all $N{:}M$ sparsity levels supported by a given accelerator but also the expected throughput improvement, our methodology can reap the benefits of accelerators supporting mixed sparsity by trading off negligible accuracy loss with both memory usage and inference time reduction for ViT models.
For instance, our approach achieves a noteworthy 2.9$\times$ reduction in FLOPs for both Swin-B and DeiT-B with only a marginal degradation of accuracy on ImageNet. Our code will be released upon paper acceptance.
\end{abstract}    
\section{Introduction}
In recent years, transformer-based neural networks have been modified for artificial intelligence tasks that include not only natural language processing but also computer vision, such as image classification \cite{VIT_dosovitskiy2021an, DEIT_pmlr-v139-touvron21a, SWIN_Liu_2021_ICCV} and object detection \cite{DETR_carion2020end, DDETR_zhu2021deformable}. Consisting of a series of transformer blocks that can effectively capture dependencies between patches in a given image, transformer-based neural networks demonstrate outstanding performance on vision tasks and replace convolutional neural networks (CNNs) as the state-of-the-art. 
For example, DeiT-B can achieve a Top-1 accuracy of 81.8\% on the ImageNet dataset by utilizing 12 transformer blocks and 17.6G parameters.
However, it is challenging to deploy such huge models on smartphones or embedded devices, given their limited memory budget and computational resources.

Various model compression methods have been proposed to reduce the requirements of memory usage and computational cost for model inference, such as quantization \cite{PTQ4ViT_DBLP:conf/nips/LiuWHZMG21, FQViT_Lin_2021, yuan2022ptq4vit} and pruning/sparsifying \cite{han2015learning, WDPViT_yu2022width, SPViT_10.1007/978-3-031-20083-0_37, DBLP:journals/corr/abs-2204-07154, SViTE}. To maintain the application accuracy, unstructured pruning methods \cite{han2015learning, OBS, ER, ERK_evci2020rigging} have been presented to remove neurons or connections from a deep neural network (DNN) that do not significantly impact accuracy and only retain important weights for computation. On the other hand, the remaining data in the weight matrix are irregularly distributed and require a high cost of encoding/indexing which induces overhead when these compressed data are sent from memory to processing units. In contrast, structured pruning methods \cite{16HEADS_NEURIPS2019_2c601ad9, WDPViT_yu2022width, SPViT_10.1007/978-3-031-20083-0_37} remove rows, columns, or channels that do not significantly impact the accuracy. Although structured pruning methods have little or no memory overhead on data encoding, the application accuracy decreases dramatically when a larger compression ratio is applied for reducing the model size. To overcome the problems mentioned above, fine-grained structured pruning (also called semi-structured pruning) has been introduced for better trade-off among the model size, compression overhead and application accuracy.

Utilizing $N{:}M$ sparsity is a type of fine-grained structured pruning method which splits every contiguous $M$ data chunks into a group of which only $N$ out of the $M$ in each group are kept for computation (\textit{i.e.,} the other $M{-}N$ data in the group are pruned) \cite{ASP, DOMINO_sun2021dominosearch, SRSTE_zhou2021learning, LBC}. Resulting sparse data representations have a lower cost on data encoding (each remaining data only needs $log_2 M$ bits for the index). Compared to unstructured pruning, $N{:}M$ sparsity is considered to be a hardware-friendly method for model compression. 

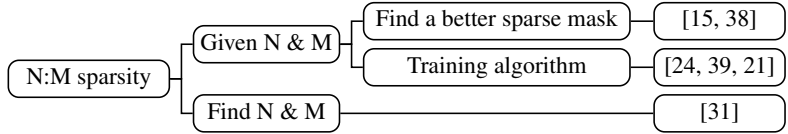
\begin{figure}[t]
    \centering
    \begin{forest}
        for tree={
          draw, semithick, rounded corners,
          fill=white, edge={draw, semithick},
          text badly centered,
        parent anchor = east, 
         child anchor = west,
                grow' = east,
                align = center,
            inner sep = 2pt,
                s sep = 1mm,    
                l sep = 3mm,    
            if n children=0{tier=last}{},
                edge path={
                    \noexpand\path [draw, \forestoption{edge}] (!u.parent anchor) -- +(1.5mm,0) |- (.child anchor)\forestoption{edge label};
                },
                if={isodd(n_children())}{
                    for children={
                        if={equal(n,(n_children("!u")+1)/2)}{calign with current}{}
                    }
                }{},
                font=\footnotesize
        }
    [N:M sparsity, text width = 20mm
        [Given N \& M, text width = 18mm
            [Find a better sparse mask, text width = 34mm
                [\cite{CAP, LBC}, text width = 16mm]
            ]
            [Training algorithm, text width = 34mm
                [\cite{ASP,SRSTE_zhou2021learning,STEP}, text width = 16mm]
            ]
        ]
        [Find N \& M, text width = 18mm
            [\cite{DOMINO_sun2021dominosearch},text width = 16mm]
        ]
    ]
    \end{forest}
    \vspace{-5pt}
    \caption{Three categories of methodologies for $N{:}M$ semi-structured pruning}
    \vspace{-15pt}
    \label{fig:NMMC}
\end{figure}

As illustrated in Fig.~\ref{fig:NMMC}, methodologies for converting a DNN to its $N{:}M$ sparse counterpart can be classified into three categories: \textbf{(1)} Methods focusing on finding a better sparse mask with a given $N{:}M$, such as CAP \cite{CAP} and LBC \cite{LBC}, precisely estimate the importance (or the sensitivity to pruning) of each weight and decide the mask based on the importance score; \textbf{(2)} Methods that obtain the target $N{:}M$ sparse network through sparse training algorithms or strategies, such as ASP~\cite{ASP}, SR-STE \cite{SRSTE_zhou2021learning}, and STEP \cite{STEP}, are proposed  to reduce the accuracy loss induced by fine-grained structured pruning by retraining the model or updating the weights; \textbf{(3)} Different from the previous two categories, which typically enforce a given/fixed sparsity level across all layers in the whole neural network, \textit{i.e.}, uniform sparsity, methods in the third category find the best $N{:}M$ sparse configuration for different layers in the network, \textit{i.e.}, layer-wise $N{:}M$ sparsity. To the best of our knowledge, DominoSearch \cite{DOMINO_sun2021dominosearch} is the only methodology employing $N{:}M$ layer-wise sparsity currently.

However, very few methods have been designed for obtaining a layer-wise customized $N{:}M$ sparse configuration for vision transformers (ViTs). Most of the methods targeting layer-wise sparsity for CNNs decide the compression ratio or the sparsity setting by considering the number of neurons or parameters in each layer \cite{ER, ERK_evci2020rigging,DOMINO_sun2021dominosearch}. Those methods compress the layers containing more parameters with higher sparsity; in contrast, the layers with fewer parameters will have lower resulting sparsity. Because ViTs usually consist of transformer blocks involving the same number of parameters, deciding the layer-wise $N{:}M$ sparse configuration according to the number of parameters is likely to have limited impact on ViTs.

To this end, we propose ELSA, a sparsity exploration framework for exploiting layer-wise $N{:}M$ sparsity on accelerating ViTs. Considering not only all $N{:}M$ sparsity levels supported by a given accelerator but also the expected throughput improvement, our methodology can reap the benefits of accelerators supporting mixed sparsity by trading off negligible accuracy loss with both memory usage and inference time reduction for ViT models. Additionally, our approach can obtain multiple target models with varying compression ratios through a single training process.

Our contributions are summarized as follows:
\begin{itemize}
    \item To the best of our knowledge, this is the first work exploring layer-wise $N{:}M$ sparse configurations for the linear modules (matrices in linear projection and multi-layer perceptron layers) in vision transformers. 
    \item Our proposed methodology can consider the $N{:}M$ sparsity levels supported by a given accelerator and search layer-wise $N{:}M$ sparse neural networks with high accuracy.
    On the other hand, the sparse configurations obtained by our method can provide an insight about the practicality of different $N{:}M$ sparsity levels for hardware designers to build accelerators supporting mixed $N{:}M$ sparsity.
    \item Considering not only the application accuracy but also the hardware efficiency, our methodology can obtain multiple layer-wise sparse configurations with high accuracy and a significant reduction in FLOPs, which is highly correlated to the throughput improvement on mixed $N{:}M$ sparsity-supporting accelerators. For instance, our approach achieves a noteworthy 2.9$\times$ reduction in FLOPs to DeiT-B with only a marginal degradation of accuracy on ImageNet classification.
\end{itemize}

\section{Preliminaries and Related Work}
\label{Sec:Preliminaries}

\begin{wrapfigure}{r}[0cm]{0.54\textwidth}
    \centering
    \vspace{-12pt}
    \includegraphics[width=1.0\linewidth]{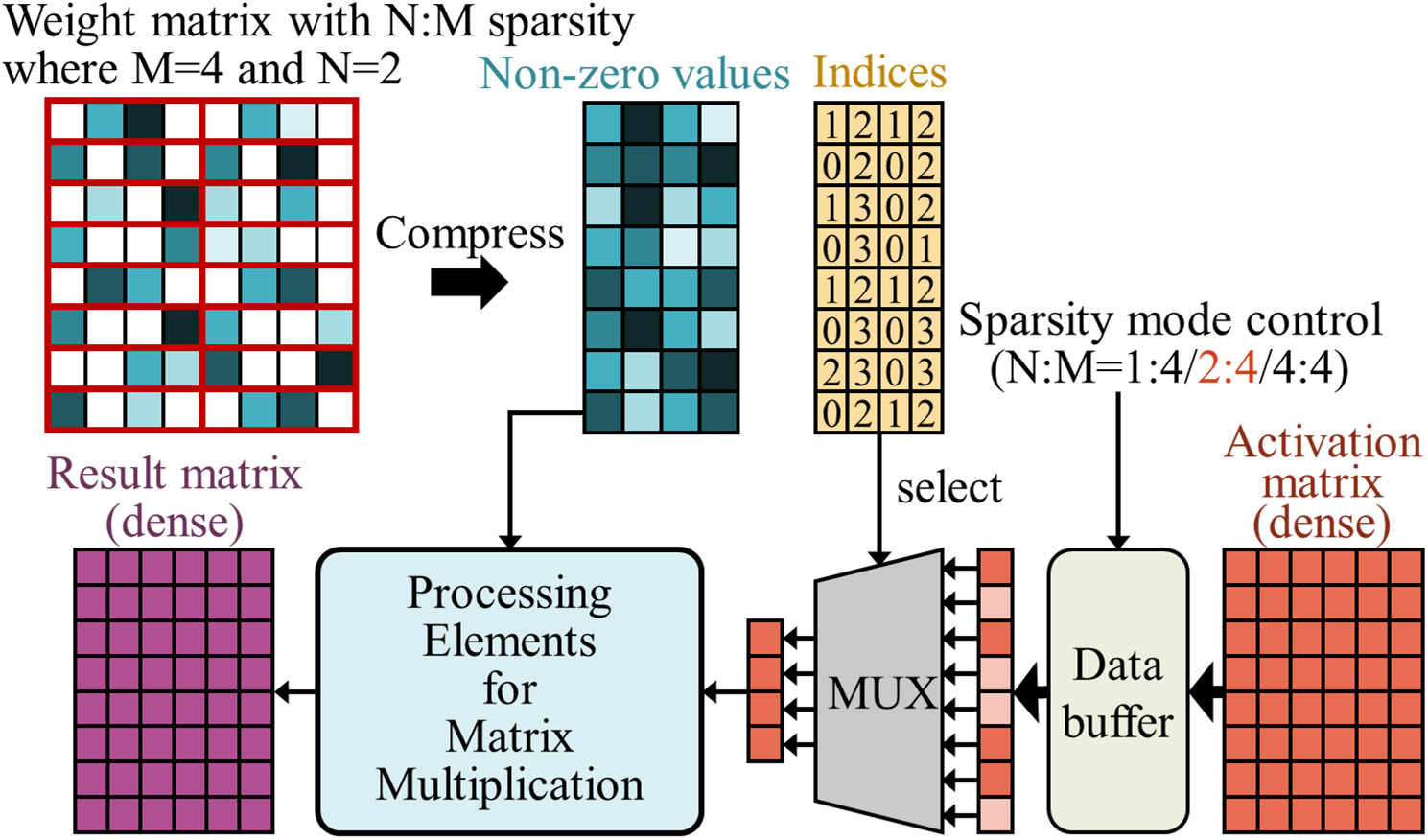}
    \vspace{-15pt}
    \caption{Accelerator for mixed sparsity where weight matrix are pruned by $N{:}M$ semi-structured pruning}
    \vspace{-15pt}
    \label{fig:NMA}
\end{wrapfigure}

To support DNN models compressed via $N{:}M$ sparsity, several accelerators have been presented \cite{A100, NMAccelerator_tvlsi22_9857911, S2TA_base8_9773187, VEGETA_base4_10071058, FIGS}. For example, the Sparse Tensor Cores in NVIDIA Ampere GPU can support 2:4 sparsity which allows a DNN model to halve its parameter count and achieve a theoretical 2$\times$ speedup on the operations of compressed 2:4 sparse matrices \cite{A100}.
In addition, the architectures presented in S2TA \cite{S2TA_base8_9773187} and VEGETA \cite{VEGETA_base4_10071058} support 4:8 weight sparsity and configurable $N$:4 sparsity in their systolic array-based accelerators, respectively, to yield speedup and energy efficiency.


Fig.~\ref{fig:NMA} illustrates a conceptual view of a configurable accelerator with $N{:}M$ sparsity support. The upper part of Fig.~\ref{fig:NMA} shows a weight matrix compressed by 2:4 sparsity. The white grids in the matrix denote values that are less significant in their 4-element group, and thus are pruned to zero. With 2:4 sparsity being applied, only the most significant two weights in the group will be retained after compressing the matrix into sparse format, where non-zeros and their 2-bit indices (for distinguishing the original position of the non-zeros from the 4-element group in $N$:4 sparsity) are stored. Therefore, not only the remaining weights but also the overhead of the corresponding data for indexing can be reduced, thereby making $N{:}M$ sparsity more hardware-friendly than unstructured pruning.

The bottom part of Fig.~\ref{fig:NMA} shows how to produce the accurate result of matrix multiplication from a sparse weight matrix and a dense activation matrix. By decoding the indices of the remaining $N$ non-zeros from each original $M$-element group in the weight matrix, the corresponding activation/input data for matrix multiplication can be selected and loaded into the processing units. Compared to the corresponding dense matrix multiplication, when the sparsity level of the current weight matrix is 2:4, only 50\% of the multiply-and-accumulate operations in the matrix multiplication will be performed. With indices of sparse matrix, control signals, associate MUXs and processing elements for sparse matrix multiplication, accelerators supporting $N{:}M$ sparsity can effectively accelerate matrix multiplication. 

If an accelerator supporting various levels of $N{:}M$ sparsity (\textit{e.g.,} $N$:4 sparsity) became available, it would be possible to obtain higher throughput by using customized, layer-wise $N{:}M$ sparsity without significant accuracy loss (compared to aggressively applying uniform sparsity, \textit{e.g.,} 1:4 sparsity). In this scenario, an effective method for finding a layer-wise sparsity setting for different DNNs on the accelerator becomes a necessity.

However, it is challenging to apply a suitable sparse configuration to ViT models so as to achieve speedups from mixed sparsity-supporting accelerators with negligible accuracy loss. For example, when a ViT model including 48 weight matrices is served on an accelerator supporting 1:4, 2:4 sparse matrix multiplication, and 4:4 dense matrix multiplication, there will be $\text{3}^\text{48}$ combinations for the resulting sparsity configurations. Furthermore, 
when a higher compression ratio is applied to accelerate operations, some essential values in the matrix will likely be pruned, which causes non-negligible accuracy loss. Accordingly, mitigating this accuracy loss necessitates extensive fine-tuning over a substantial number of epochs to maintain the accuracy of the compressed model. Therefore, an efficient methodology is required for determining a suitable layer-wise sparse configuration among the many possible combinations without resorting to exhaustive search and fine-tuning.

\section{Methodology}
\label{sec:method}
\paragraph{Problem Formulation}

\begin{wrapfigure}{r}[0cm]{0.4\textwidth}
    \centering
    \vspace{-17pt}
    \includegraphics[width=1.0\linewidth]{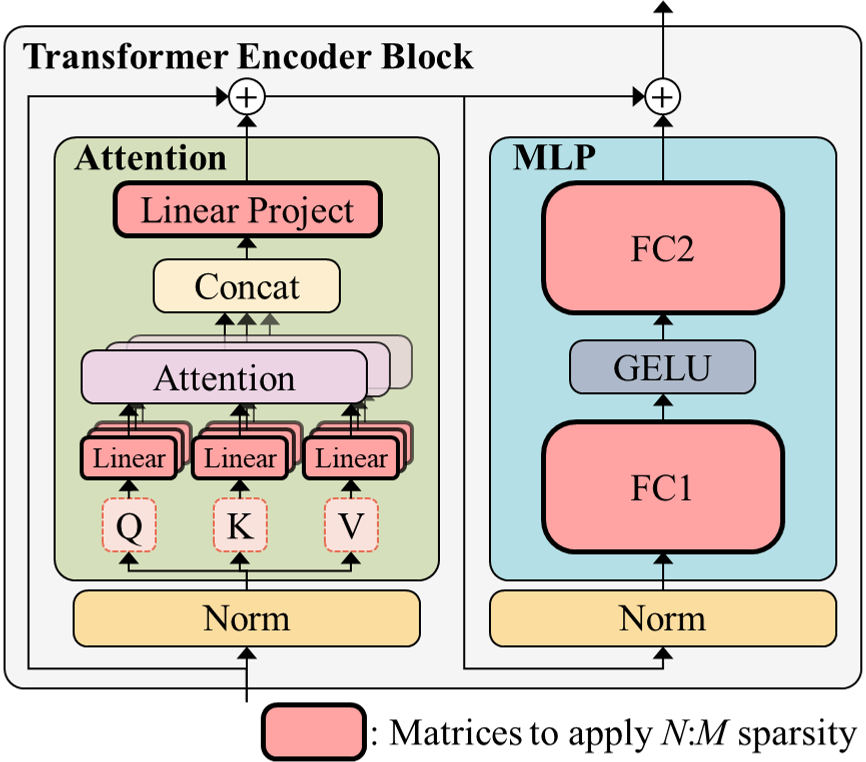}
    \vspace{-18pt}
    \caption{A classic transformer block and the matrices to be sparsified in our methodology}
    \vspace{-15pt}
    \label{fig:tblock}
\end{wrapfigure}

Assuming the availability of an accelerator supporting mixed $N{:}M$ sparsity levels for matrix multiplication operations, one can apply the $N{:}M$ semi-structured pruning to the linear modules of ViTs in the pursuit of computational efficiency. Consider a pretrained classic ViT comprising $B$ transformer blocks, each including four linear modules, as depicted in Fig.~\ref{fig:tblock}. Assuming the accelerator supports a set of $K$ different sparsity levels $\mathcal{S}$ = $\bigl\{(N_1{:}M_1),\dots,(N_K{:}M_K)\bigr\}$, our objective is to determine a sparse configuration (\textit{i.e.,} a sequence of sparsity levels) denoted as $\alpha$~=~$\bigl(s^{1},\dots,s^{L}\bigr)$, where $s^{l}\in\mathcal{S}$ and $L$ = $4B$. This configuration $\alpha$ is then used to convert the pretrained weight values, represented as $\mathcal{W}$ = $\bigl(W^{1},\dots,W^{(L)}\bigr)$ from a dense to a sparse representation. To harness the acceleration potential of the sparsity-supporting accelerator while minimizing associated performance degradation, it is essential to determine the optimal sparse configuration.

\paragraph{Challenge} Deploying a ViT model on accelerators supporting flexible sparse matrix multiplication typically involves two steps: (1) determining the $N{:}M$ sparsity level for each matrix in the model, followed by (2) fine-tuning the model to mitigate the accuracy loss resulting from sparsification. However, deciding the layer-wise sparsity configurations for ViTs can be more challenging compared to convolution-based models, primarily because ViTs comprise multiple transformer blocks involving the same number of parameters. Consequently, previous heuristic-based approaches to sparsity level selection, as previously employed in the context of convolutional models \cite{ER, Lamb, ERK_evci2020rigging}, confront difficulties in selecting configurations that offer both superior accuracy and substantial computational speed-ups on ViTs.

Furthermore, the process of sparsification often results in a significant degradation of accuracy, especially when matrices are pruned in a more structured manner (\textit{i.e.,} through structured or semi-structured pruning). This necessitates extensive fine-tuning over a substantial number of epochs to restore the application accuracy of the given model, enabling a precise evaluation of the chosen sparse configurations.

To address the significant fine-tuning demands, we adopt the supernet paradigm~\cite{SPOS, Single_path_NAS, OFA} which was originally designed to avoid training each candidate network from scratch to accurately evaluate the corresponding performance while searching network architecture hyper-parameters, \textit{e.g.}, the kernel size and the number of channels for CNNs. Through customized encoding strategies, all candidate networks within the search space, referred to as subnets, are encoded into an over-parameterized supernet. These subnets are then trained jointly, effectively amortizing the fine-tuning requirements for evaluating each specific/chosen combination of architecture hyper-parameters. By training only once, we can yield the network corresponding to any specific network configuration by directly inheriting from the supernet without exhaustively training each of them one by one.

In this work, we propose to extend these principles to address the exploration complexities of obtaining suitable layer-wise $N{:}M$ sparse networks. Specifically, we design a supernet construction scheme aimed at incorporating all candidate $N{:}M$ sparse networks, each corresponding to an $N{:}M$ sparse configuration $\alpha$ within the search space \mbox{$\mathcal{A}$ = $\underbrace{\mathcal{S} \times \mathcal{S}\times\dots\times\mathcal{S}}_{L}$}.
\vspace{-0.5ex}
By harnessing this supernet, we can directly derive specific sparse networks without the need for additional fine-tuning. This not only facilitates evaluation of specific sparse configurations $\alpha$ but also expedites their deployment on hardware accelerators to enhance computational throughput. 

Detailed description of the algorithms for our supernet construction, training, and the search processing for deriving sparse subnets from our supernet will be provided in the subsequent subsections.




\subsection{Supernet Construction -- All Sparsity Configurations in One Supernet}


We initiate our layer-wise sparsity exploration by first constructing a supernet, which involves the determination of all layers that are utilized to perform matrix multiplication in a ViT model, such as the linear projection and multi-layer perceptron layers. For example, DeiT\nobreakdash-S consists of 12 transformer blocks, each containing 4 target layers (\textit{i.e.,} qkv, linear projection, fc1 and fc2); thus, the sparsity configuration of the 48 layers should be determined before deploying a sparse DeiT\nobreakdash-S model on an accelerator supporting mixed sparsity. Subsequently, 
all supported sparsity levels of a given accelerator are encoded as possible choices available to each layer. If an accelerator offers 1:4, 2:4, and 4:4 sparsity options, each target layer within our supernet will contain 3 sparsity choices, where one of them will be chosen at a time, and the corresponding sparsity level will be applied for sparse matrix multiplication when the layer is computed. Moreover,
each combination of sparsity choices for a given ViT model is regarded as a sparse configuration, which can be utilized to derive a sparse subnet within our supernet. After constructing a supernet that contains all sparsity configurations, we train the supernet before searching for a suitable configuration.


\paragraph{Subset-Superset Relationship}

\begin{figure}[!htb]
   \begin{minipage}{0.54\textwidth}
     \centering
     \includegraphics[width=1.\linewidth]{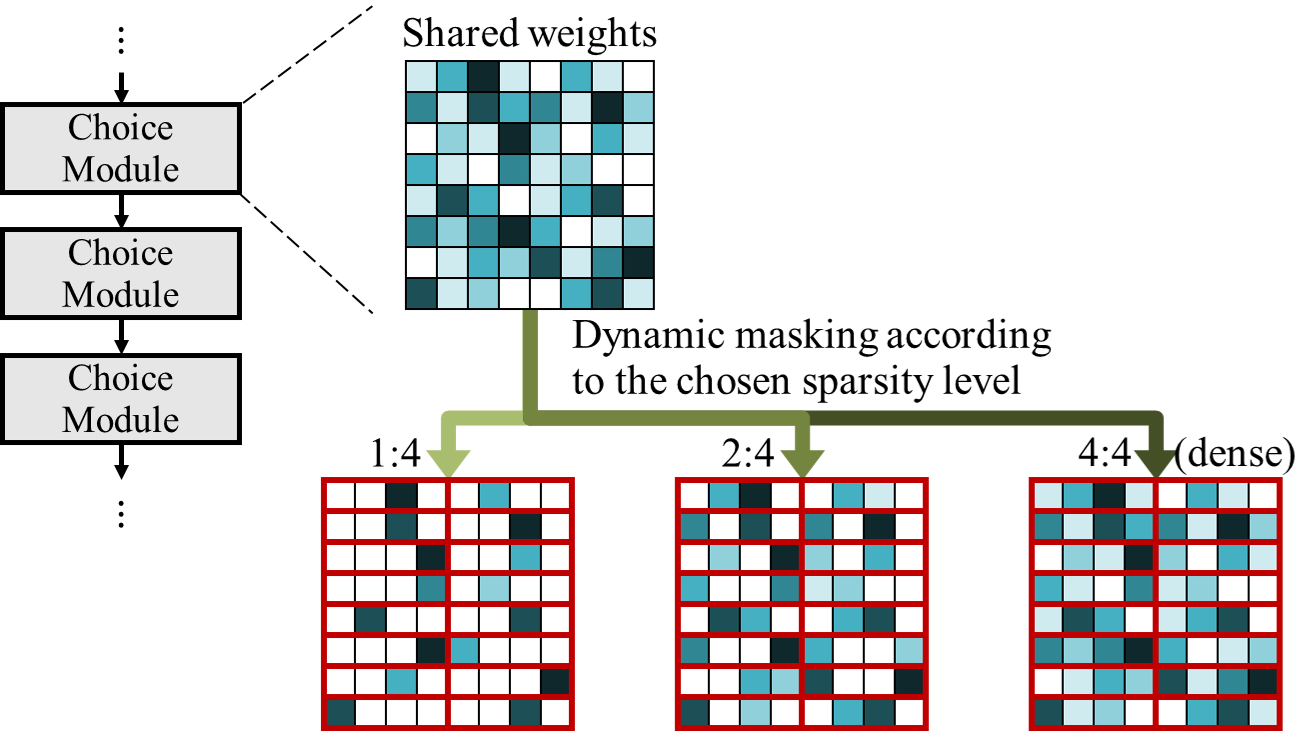}
     \vspace{-15pt}
     \caption{Shared weight values and dynamic masking for each layer in a transformer-based model (including all linear projection and multi-layer perceptron layers)}
     \label{fig:dm}
   \end{minipage}\hfill
   \begin{minipage}{0.43\textwidth}
     \centering
     \includegraphics[width=1.\linewidth]{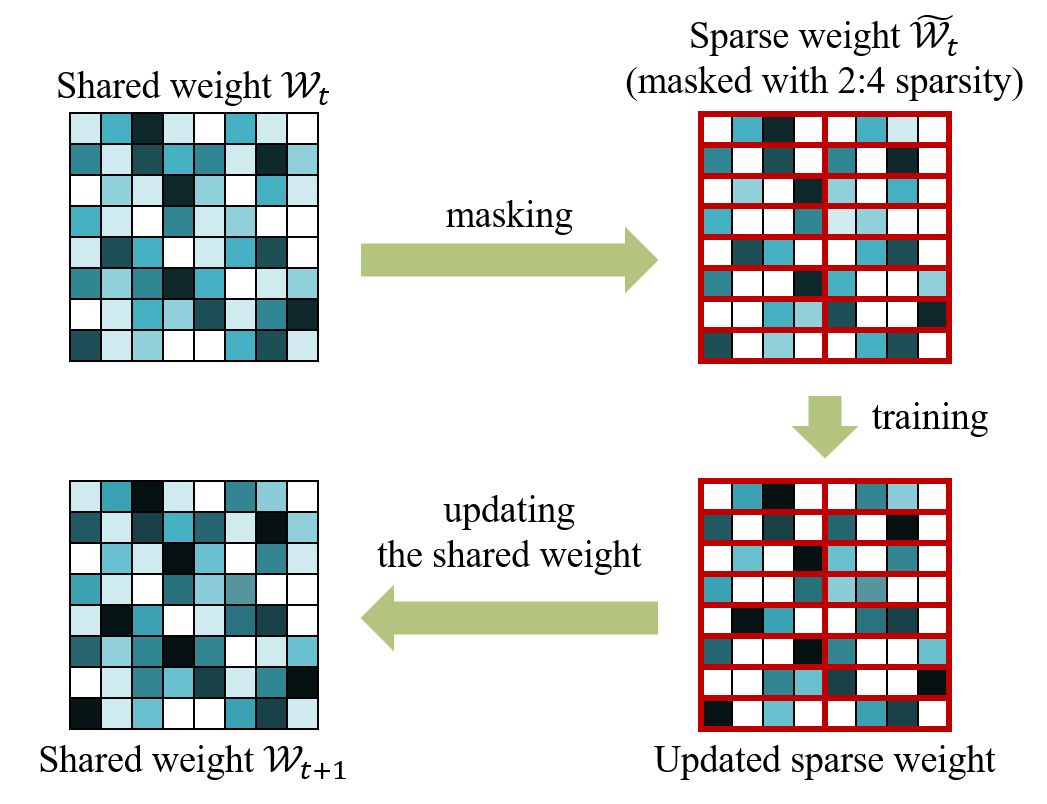}
     \vspace{-15pt}
     \caption{Example of updating the shared weight with 2:4 sparsity being applied}
     \label{fig:wupdate}
   \end{minipage}
\end{figure}

We notice the subset-superset relationship among different $N$:$M$ sparsity levels and utilize it to construct the supernet. 
Specifically, the non-zeros selected under higher $N{:}M$ sparsity level form a subset of those selected under lower ones when the same criteria for pruning is applied. For the $N$:4 sparsity levels illustrated in Fig.~\ref{fig:dm}, the cells of darker color mean their values are more significant; both 1:4 and 2:4 sparsities select their non-zeros according to the significance of those weights. It can be seen that the weight selected when 1:4 sparsity is applied must be also selected when 2:4 sparsity is applied. Furthermore, those weights are all included
in the dense matrix (\textit{i.e.,} using 4:4 sparsity).

By leveraging this relationship, instead of building up a supernet that has different instances of sparse weights for each sparsity level in each matrix and thus causing huge memory overhead for the supernet training, we only maintain a stack of weight matrices $\mathcal{W}_\mathcal{A}$ = $\bigl(W_{sup}^1,\dots, W_{sup}^L\bigr)$, where $W_{sup}^i$ denotes the shared weight matrix for the $i^\text{th}$ linear module. That is, all $N$:$M$ sparsity choices share the same weight matrix in each layer within our supernet.

\paragraph{Dynamic Masking} During the process of both training and evaluation, given a sparse configuration $\alpha$ = $\bigl(s^1,\dots,s^{L}\bigr)$,
we generate the sparse networks $W_{\mathcal{A}}(\alpha)$ = $\bigl(W^{1}_{sup}(s^1),\dots,W^{L}_{sup}(s^L)\bigr)$ by dynamically projecting the shared parameters ${W^l_{sup}}$ of $l^\text{th}$ linear modules into corresponding sparse matrix \mbox{$W^{l}_{sup}(s^l)$} according to the specific $N{:}M$ sparsity level $s^l$. To achieve this, we treat $M$ consecutive parameters as a group ${G}$ = $\{w_1, \dots,w_M\}$, where $G \subset W^l_{sup}$. We prune the $M-N$ parameters that have the least significant saliency score $\rho(\cdot)$, \textit{i.e.,} the importance score for each weight. As Fig.~\ref{fig:dm} illustrates, when 1:4 sparsity is chosen, three parameters in each 4-element group will be pruned.
Mathematically, we define $\zeta$ as the $N^\text{th}$ smallest saliency score among $G$, and the mask value $m_i$ for weight elements in a group can be calculated as:
\begin{equation}
    m_i=\begin{cases} 1, &\textrm{if } \rho(w_i) \geq \zeta\\
    0, &\textrm{if } \rho(w_i) < \zeta
    \end{cases},\;\text{for } i \in \{1,\dots ,M\}
\label{eq:mask}
\end{equation}

With binary mask $m$, we can determine the sparse weights by pruning the parameter $w_i \in G$ to zero if the corresponding mask value $m_i$ is equal to 0. In our framework, the saliency score function $\rho(\cdot)$ can be a variety of metrics, including the weight magnitude and Taylor score \cite{DBLP:conf/iclr/MolchanovTKAK17, DBLP:conf/cvpr/MolchanovMTFK19}. For simplicity, we utilize weight magnitude as the default saliency score estimator.

In general, by leveraging the subset-superset relationship, all given sparsity levels can be encoded into our supernet. This approach ensures that the same number of weights in the given neural network is maintained, eliminating the need for multiple weight replicas across different sparsity levels.~This not only reduces memory cost but also significantly reduces training overhead, streamlining the computational process. Furthermore, this shared weight approach enhances the convergence of sparse subnetworks, as selecting a particular sparsity level for optimization during supernet training simultaneously trains its subsets and supersets. As illustrated in Fig.~\ref{fig:wupdate}, if 2:4 sparsity is applied, the corresponding sparse weight matrix will be trained and the shared weight will also be updated.

\subsection{Supernet Training and Subnetwork Sampling}
\label{sec:supernet_training}
Supernet training aims to comprehensively train all sparse networks within a designated search space simultaneously. The essence of this process is captured by the following objective function:
\begin{equation}
\label{SPOS training}
     \mathcal{W}_{\mathcal{A}}^{*} =  \underset{\mathcal{W}_{\mathcal{A}}}{\text{argmin}}\mathbb{E}_{a \thicksim \Gamma(\mathcal{A})} [ \mathcal{L} (\alpha, \mathcal{W}_{\mathcal{A}}(\alpha) ],
\end{equation}
Here, $\mathcal{W}_{\mathcal{A}}$ represents the weights of the supernet, $\mathcal{L}$ denotes the training loss. Each sparse configuration is represented by $\alpha \in \mathcal{A}$, where $\mathcal{W}_{\mathcal{A}}(\alpha)$ denotes the sparse weights corresponding to each configuration $\alpha$. The objective corresponds to minimizing the expected loss across all existing sparse subnetworks within the search space, $\mathcal{A}$.  In each optimization step, a sparse architecture is sampled from a prior distribution $\Gamma(\mathcal{A})$. Such sparse architectures are trained to minimize the training loss $\mathcal{L}$ and the gradient will be updated back to the shared supernet weights.

\begin{figure}[!htb]
    \begin{minipage}{0.48\textwidth}
        \centering
        \includegraphics[width=1.\linewidth]{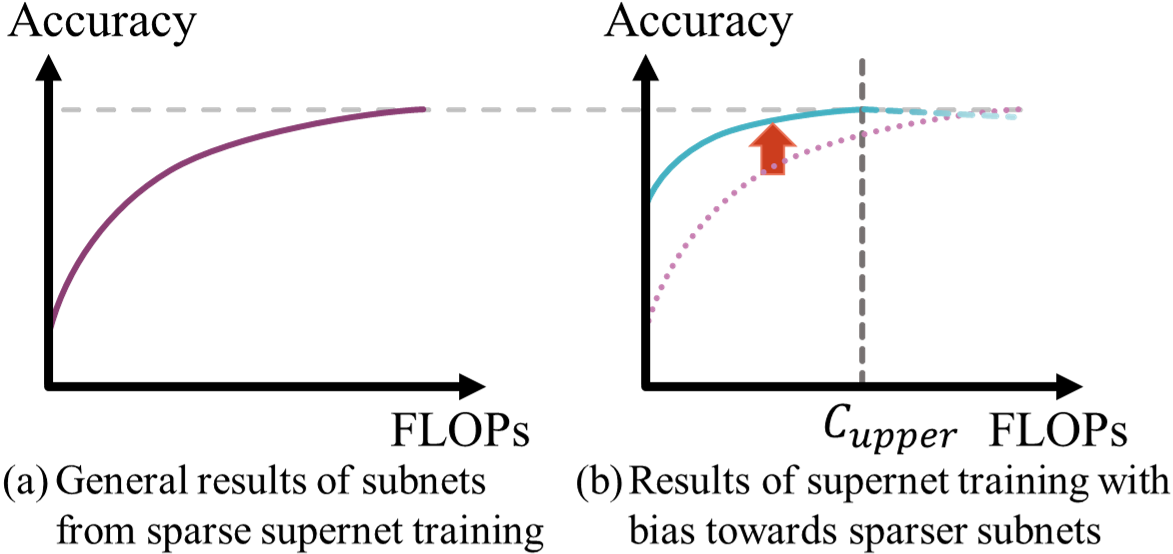}
        \vspace{-15pt}
        \caption{Supernet training with a bias towards sparser subnets}
        \label{fig:sup_training_bias}
   \end{minipage}\hfill
   \begin{minipage}{0.5\textwidth}
        \centering
        \includegraphics[width=1.\linewidth]{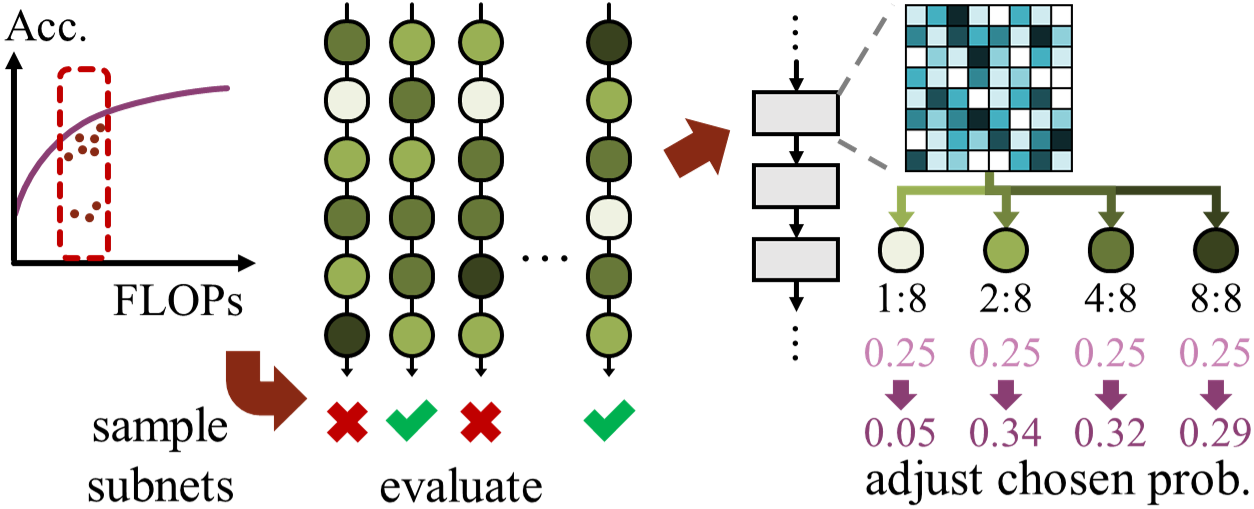}
        \vspace{-16pt}
        \caption{Sample subnets and automatically lower the probabilities of choices that are too sparse to maintain accuracy}
        \label{fig:werewolf}
   \end{minipage}
\end{figure}

\paragraph{Prioritizing Sparser Subnets}  In supernet training, managing the complexity of the search space is important. We streamline this process by strategically focusing on (i) configurations that can be converted into sparser networks for obtaining further throughput improvement or (ii) configurations within a user-defined threshold of computational cost, $C_{upper}$. In practical terms, this means that only configurations with a computation cost $\mathcal{F}(\alpha)$ less than or equal to $C_{upper}$ are included in the sampling process.
This tailored adjustment enables the search space to adapt to various practical constraints, such as hardware limitations, making the search more aligned with available computational resources and real-world applicability. In addition, by considering a given $C_{upper}$ value, our supernet training can focus on sparser subnets, which usually require more effort to recover model accuracy. As shown in Fig.~\ref{fig:sup_training_bias}, the performance of sparser subnets will be higher with $C_{upper}$ being considered due to the prioritization during the supernet training.

\paragraph{Sampling Strategy} Before we delve into our proposed prior distribution for candidate configuration, we first review the vanilla approach. In the conventional method \cite{SPOS}, all candidate configurations are treated equally. Specifically, sparse configurations are sampled by uniformly selecting the sparsity level $s_i$ from $\mathcal{S}$ for each $i^\text{th}$ linear module independently.

Leveraging insights from the central limit theorem \cite{CLT}, it becomes evident that under this vanilla sampling paradigm, the distribution of computational cost $\mathcal{F}(\alpha)$ will approximately converge to a normal distribution. Given the architecture of our weight-sharing supernet, when a particular sparse configuration $\alpha$ is sampled, similar configurations with nearly identical computational costs are also indirectly trained. This biases the training towards architectures with computational costs clustering around the mean, leaving architectures with extremal computational costs, especially those with higher sparsity, subject to undertraining.

To motivate and ensure a more balanced exploration across the architectural spectrum, we employ a two-step sampling strategy. This process starts by discretizing the computational costs into distinct levels, represented as $C_1, C_2, \ldots, C_K$, where $C_1$ corresponds to the minimum computational cost ($C_{lower}$), and $C_K$ corresponds to the user-specified maximum computational cost ($C_{upper}$). Mathematically, this discretization can be defined by denoting a set of intervals $\mathcal{I}$~=~$\{[C_i, C_{i+1})_{1\leq i \leq K-1}\}$. With these intervals, the sampling unfolds in two stages: first, a computational cost interval $I$~=~$[C_i, C_{i+1})$ is chosen uniformly from~$\mathcal{I}$, and then, within this selected interval, an architecture is uniformly sampled. Based on this strategic approach, we can ensure the balanced training for architectures of different computational costs, thereby alleviating the undertraining issue for the extremal architectures of high sparsity.

\begin{wrapfigure}{R}{0.49\textwidth}
\vspace{-5ex}
\begin{minipage}{1.0\linewidth}
    \begin{algorithm}[H]
        \caption{Update Chosen Probabilities}
        \label{algo:auto_ch_filter}
        \footnotesize
        \textbf{Input:} 
        \begin{minipage}[t]{0.9\linewidth}
            current trained supernet weight $\mathcal{W}_{\mathcal{A},t}$, \\
            current probabilities for sparsity choices $\mathcal{P}_t$, \\
            evaluation data $D_{eval}$, \# of subnets $N_{eval}$,\\
            accuracy threshold $Acc_{th}$
        \end{minipage} \\
        \textbf{Output:} 
        \begin{minipage}[t]{0.8\linewidth}
            updated probabilities $\mathcal{P}_{t+1}$
        \end{minipage}
        \begin{algorithmic}[1]
            \State Init choice scores $\mathcal{H}$ with zeros
            \For{$n$ = $1,\dots,N_{eval}$}
                \State Choose a configuration $\alpha$ based on $\mathcal{P}_t$
                \State Derive the sparse weights $\mathcal{W}_{\mathcal{A},t}(\alpha)$ from $\mathcal{W}_{\mathcal{A},t}$
                \Statex \hspace{\algorithmicindent} $\mathcal{W}_{\mathcal{A},t}(\alpha)$ = $\bigl(W^{l}_{sup, t}(s^1),\dots,W^{l}_{sup, t}(s^L)\bigr)$
                \State Evaluate the subnet with $D_{eval}$
                \For{$l$ = $1,\dots,L$}
                    \State Update $h_{l,s^l}$ with the evaluation score
                \EndFor
            \EndFor
            \For{each element $h_{l,s_i} \in \mathcal{H}$}
                \If{$h_{l,s_i} < Acc_{th}$}
                    \State $h_{l,s_i} = 0$
                \EndIf
            \EndFor
            \For{$l$ = $1,\dots,L$}
                \State $p_{l,s_i}$ = $h_{l,s_i} / \sum_{s_i \in \mathcal{S}}h_{l,s_i}$
                \Statex \hspace{\algorithmicindent} $/*$ Ensure $0 \leq p_{l,s_i} \leq 1$ \Statex \hspace{\algorithmicindent} \hspace{2.5ex} and $\sum_{s_i \in \mathcal{S}}p_{l,s_i} = 1$ for each layer $*/$ 
            \EndFor
            \State return probability $\mathcal{P}_{t+1}$
            \vspace{-0.3ex}
        \end{algorithmic}
    \end{algorithm}
\end{minipage}
\vspace{-1ex}
\end{wrapfigure}

\normalfont

\paragraph{Automatic Choice Filtering}\label{para:acf} When exploring further sparsity by providing a search space with much sparser choices, such as 1:8 or 1:16 sparsity, the supernet training might be negatively impacted because those choices are too sparse for some layers to propagate information into the subsequent layers. Therefore, we propose an automatic choice filtering method to remove sparsity choices that are too sparse to maintain accuracy in some layers during the training stage. To filter unsuitable sparsity choices out from the search space, we maintain the probabilities $\mathcal{P}$ of sparsity choices to be chosen through the process described in Algorithm~\ref{algo:auto_ch_filter}. 
Here, every layer $l$ has a set of sparsity choices $\mathcal{S}$ and each sparsity choice $s_i$ has an associated probability $p_{l, s_i}$ where $p_{l,s_i} \geq 0$ and $\sum_{s_i \in \mathcal{S}}p_{l,s_i}$ = $1$. After several training epochs, sparse subnets are randomly sampled from each interval $I$ 
and their accuracies evaluated with a proxy dataset (a subset of the validation dataset to reduce the evaluation time). The accuracy of a sampled subnet will be regarded as the evaluation score to update the probability of each sparsity level in its sparse configuration, \textit{i.e.}, all $p_{l, s^l}$ where $s^l$ is the sparsity choice of the layer $l$ in the sparse configuration $\alpha$~=~$ (s^1,\dots,s^L)$. By evaluating a certain number of subnets, $\mathcal{P}_{t+1}$ represents the influence of sparsity choices on the accuracy of different layers. When a sparsity level is chosen to sparsify a layer and induces a larger accuracy drop, its score will be low. If the score of a sparsity level is lower than the given threshold $Acc_{th}$, its probability will be 0 (not be chosen in the following training). Otherwise, the scores of the sparsity choices in each layer will be normalized and utilized as the probabilities $\mathcal{P}_{t+1}$ for the following training. For the detailed training recipe to enhance our supernet training, please see Appendix~\ref{sec:appendix_algo}.

\subsection{Searching for the Pareto-optimal Solutions}
To obtain a suitable sparse configuration from the trained supernet, we rely on evolutionary search and our proposed automatic choice filtering strategy (Algo.~\ref{algo:auto_ch_filter} in Section~\ref{para:acf}) to determine a suitable trade-off between accuracy and acceleration. Evolutionary algorithms contain four steps: initialization, mutation, crossover, and selection. First, we initialize the population of subnetworks by performing our two-step sampling strategy to sample several sparse configurations that consist of a sequence of sparsity choices for the ViT. Second, the chosen sparsity level for each layer in a subnetwork is mutated (\textit{e.g.}, the sparsity level is changed from 2:4 to 1:4) with a mutation probability. Third, we perform crossover on multiple pairs of subnetworks in the population by changing partial sparsity settings to generate new sparsity combinations. 
After that, according to a fitness function, only the sparsity setting with the top-\textit{k} best fitness score is retained in the population for the next iteration (those with lower scores are removed). 
After several iterations of mutation, crossover and selection, suitable sparse configurations will be obtained, and the corresponding high-accuracy subnetworks can be extracted from our supernet without further fine-tuning.


\section{Experiments}
\label{sec:exp}

\paragraph{Dataset and Benchmarks}
Our experiments are conducted on the ImageNet-1k dataset for image classification tasks. We apply our methodology to DeiT, which includes multiple transformer encoder blocks, and to the improved Swin-Transformer, which employs a hierarchical design and shifted window approach.
For the detailed settings of our supernet training, please see Appendix~\ref{sec:appendix_setting}.


\begin{table}[ht]
\centering
\caption{
Experimental results of the proposed ELSA methodology on various vision transformers. Accuracy denotes the Top-1 accuracy measure on the ImageNet-1K validation set.}
\def\arraystretch{1.2}
\resizebox{0.6\linewidth}{!}{
  \begin{tabular}{ l |  c  | c  |  c  }
    \ChangeRT{2pt}
    {Model} & {Sparsity Pattern} & {FLOPs} & {Accuracy}  \\
    \ChangeRT{1.75pt}
    DeiT-S                           & Dense                            &  4.7G & 79.8\% \\
    \ChangeRT{0.5pt}
    ELSA-DeiT-S-2:4                  & Uniform 2:4                      &  2.5G (1.00$\times$) & 79.1\% \\
    \ChangeRT{0.5pt}
    \multirow{2}{*}{ELSA-DeiT-S-N:4} & \multirow{2}{*}{Layer-wise N:4}  &  2.2G (1.14$\times$) & 79.0\%\\
                                     &                                  &  2.0G (1.25$\times$) & 78.3\% \\
    \ChangeRT{1.75pt}
    DeiT-B                           & Dense                            & 17.6G & 81.8\% \\
    \ChangeRT{0.5pt}
    ELSA-DeiT-B-2:4                  & Uniform 2:4                      & 9.2G (1.00$\times$) & 81.6\% \\
    \ChangeRT{0.5pt}
    \multirow{2}{*}{ELSA-DeiT-B-N:4} & \multirow{2}{*}{Layer-wise N:4}  & 7.0G (1.30$\times$) & 81.6\% \\
                                     &                                  & 6.0G (1.53$\times$) & 81.4\% \\
    \ChangeRT{1.75pt}
    Swin-S                           & Dense                            & 8.7G & 83.2\% \\
    \ChangeRT{0.5pt}
    ELSA-Swin-S-2:4                  & Uniform 2:4                      & 4.6G (1.00$\times$) & 82.8\% \\
    \ChangeRT{0.5pt}
    \multirow{2}{*}{ELSA-Swin-S-N:4} & \multirow{2}{*}{Layer-wise N:4}  & 4.0G (1.15$\times$) & 82.8\% \\
                                     &                                  & 3.5G (1.31$\times$) & 82.5\% \\
    \ChangeRT{1.75pt}
    Swin-B                           & Dense                            & 15.4G & 83.5\%  \\
    \ChangeRT{0.5pt}
    ELSA-Swin-B-2:4                  & Uniform 2:4                      & 8.0G (1.00$\times$)  & 83.1\%   \\
    \ChangeRT{0.5pt}
    \multirow{2}{*}{ELSA-Swin-B-N:4} & \multirow{2}{*}{Layer-wise N:4}  & 6.0G (1.33$\times$) & 83.0\%   \\
                                     &                                  & 5.3G (1.51$\times$) & 82.8\%   \\
    \ChangeRT{2pt}
    \end{tabular}
}
\label{Tab: Sparse Network}
\end{table}



\subsection{Pruning Results on ImageNet-1K}
In our methodology, we employ $N$ as a power-of-two to define our search space and engage in a cohesive training process of a supernet. This supernet houses various sparse subnetworks. Using evolutionary search, we strategically navigate the search space to identify the best layer-wise sparsity configurations. Our results are presented in Table~\ref{Tab: Sparse Network}. All models are directly inherited from the supernet, necessitating no additional adjustments, fine-tuning or post-training.

\begin{wrapfigure}{r}[0cm]{0.5\textwidth}
    \centering
    \vspace{-13pt}
    \includegraphics[width=1.0\linewidth]{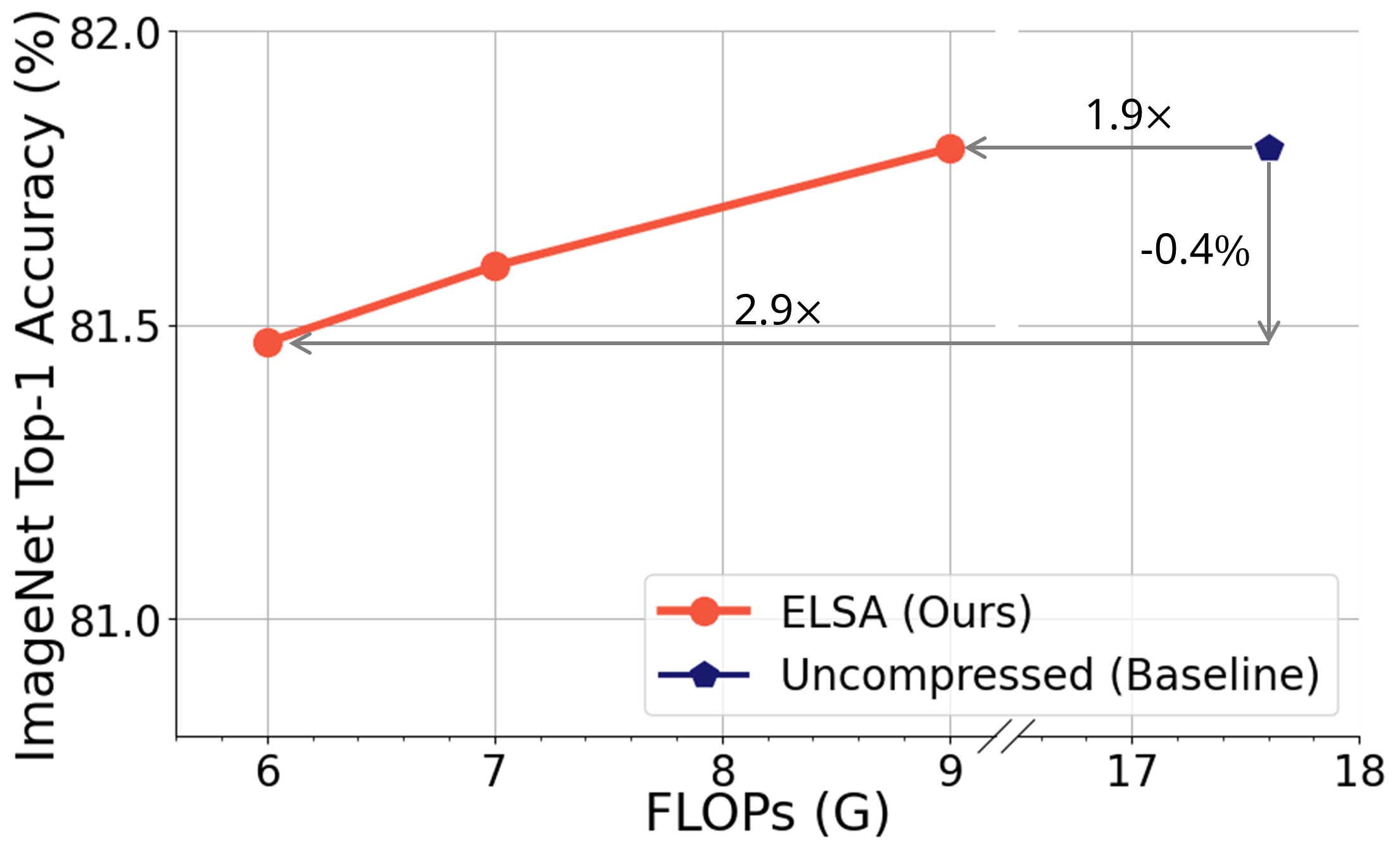}
    \vspace{-22pt}
    \caption{Result of sparsity exploration on DeiT-B}
    \vspace{-10pt}
    \label{Fig:DeiT-B Compare To Dense}
\end{wrapfigure}

\paragraph{Comparison with the Dense Model (uniform 4:4)}
As shown in Table~\ref{Tab: Sparse Network}, our approach demonstrates the capability to significantly reduce FLOPs through exploring $N{:}M$ sparsity while maintaining model performance. For instance, when applying our pruning strategy to DeiT-B, we achieve a noteworthy $2.9\times$ reduction in computational costs, with only a marginal decline in accuracy as depicted in Fig. \ref{Fig:DeiT-B Compare To Dense}. Pursuing a FLOP reduction of approximately 50\%, our employment of an evolutionary search approach has directed us towards the uniform 2:4 sparsity setting from our well-trained supernet. Furthermore, the sparse network with uniform 2:4 sparsity setting can facilitate an almost twofold reduction in computational costs with only 0.2\% reduction in accuracy. Similar efficiency is observed in the case of the Swin-Transformer models, thus confirming the robustness of our methodology.

\paragraph{Benefits of Layer-wise Sparsity}
The adoption of layer-wise sparsity serves as a foundational aspect of our methodology, presenting notable advantages in comparison to the conventional 2:4 sparse configuration. As demonstrated by the DeiT-B model, our approach, through the exploration of layer-wise sparsity, has achieved improved efficiency, manifesting a $1.53\times$ reduction in FLOPs (6.0G \textit{vs.} 9.2G). In addition, the reduction in FLOPs is highly correlated to the speedup obtained by deploying the sparsified network on $N{:}M$ sparsity-supporting accelerators. This result indicates the capacity of layer-wise sparsity to harness heightened rates of hardware acceleration, a trend that is also evident in the Swin-B model. Notably, the Swin-B model reported a $1.51\times$ enhancement in computational cost efficiency (5.3G \textit{vs.} 8G) as a direct consequence of our method.

It is worth mentioning that the FLOPs reduction yielded by our semi-structured sparsity exploration can theoretically be translated into runtime speed-ups when paired with emerging $N$:$M$ sparsity-supporting hardware accelerators. Assuming optimal hardware utilization, a network with 2:4 sparsity can attain nearly a $2\times$ speed enhancement on platforms like the NVIDIA Ampere GPU \cite{A100}. Likewise, sparse networks employing layer-wise $N$:4 sparsity, such as ELSA-DeiT-B-N:4, could expect speed-ups of up to 
$2.9\times$ when deployed on VEGETA \cite{VEGETA_base4_10071058}. Moreover, it is crucial to emphasize the adaptability of our methodology, which allows for the exploration of various semi-structured sparsity patterns. This adaptability facilitates seamless integration with different hardware designs, such as S2TA \cite{S2TA_base8_9773187}, which targets $N$:8 sparsity, aiding the identification of efficient configurations for developing well-trained sparse networks without compromising accuracy.




\subsection{Comparison to Other Sparse Configuration Search Methods}
\label{sec:exp_cmp_search}
To demonstrate the efficacy of our methodology in identifying an optimal layer-wise sparse configuration, we compare our approach to other prevailing sparsity selection paradigms. The methods under comparison are \textbf{Erd\H{o}s-R\'enyi (ER}~\cite{ER}, originally designed to determine layer-wise unstructured sparsity; we modified ER for $N{:}M$ sparsity by initially deciding the unstructured sparsity level and subsequently rounding it to the closest $N{:}M$ sparsity level\textbf{)} and \textbf{DominoSearch} \cite{DOMINO_sun2021dominosearch} (an $N{:}M$ sparsity level selection approach tailored for CNNs). For more descriptions about these two methods, please see Appendix~\ref{sec:appendix_ER_Domino}.

\begin{wrapfigure}{r}[0cm]{0.5\textwidth}
    \centering
    \vspace{-15pt}
    \includegraphics[width=1.0\linewidth]{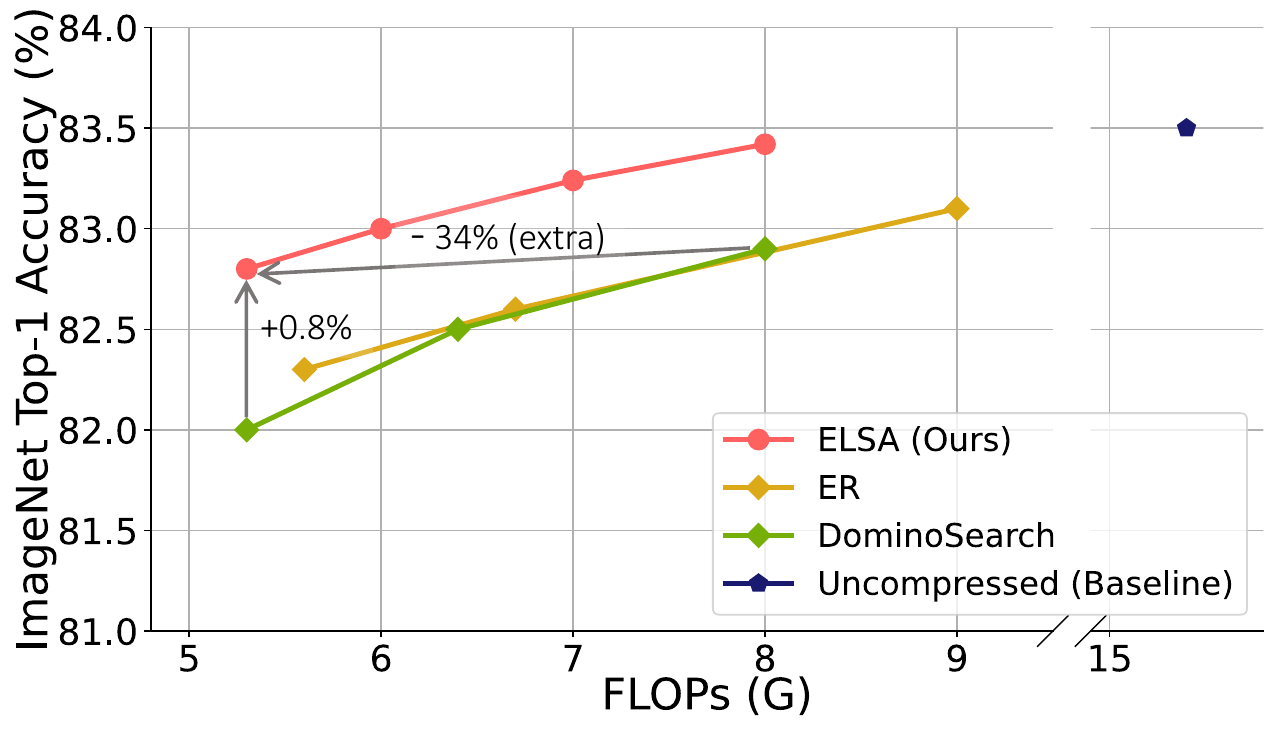}
    \vspace{-22pt}
    \caption{Comparisons of ELSA with other search methods on Swin-B}
    \vspace{-10pt}
    \label{Fig:Swin-B Compare}
\end{wrapfigure}




For our experiments, the aforementioned strategies were employed on pretrained ViTs to decide the sparsity level for each targeted linear module. Once the sparsity levels are determined, we derive the sparse sub-networks from the fully-trained supernet and evaluate their performance on the ImageNet-1k validation dataset.

From the visual representation in Fig. \ref{Fig:Swin-B Compare}, it is evident that our approach consistently outperforms the baseline methodologies on the Swin-B architecture. Our methodology exhibits a more favorable trade-off between FLOPs and accuracy. Notably, our method manages to identify a sparsity configuration that achieves a significant extra 34\% reduction in FLOPs compared to the DominoSearch strategy, while still maintaining comparable, if not superior, accuracy. Additionally, an improvement of 0.8\% in Top-1 accuracy is observed when aligning the FLOPs. Lastly, the plots emphasize that our results consistently reside on the Pareto frontier, indicating a better balance between computational efficiency and performance. This visual evidence underscores the effectiveness of our proposed layer-wise sparsity exploration technique in ViTs. For a detailed visualization of the sparse configurations searched by the discussed algorithms, please see Appendix~\ref{sec:appendix_visualization}.


\section{Conclusion}
This paper presents ELSA, a first-of-its-kind layer-wise $N{:}M$ sparsity exploration methodology for accelerating ViTs. We address the challenge of selecting suitable layer-wise sparse configurations for ViTs on $N{:}M$ sparsity-supporting accelerators. Leveraging the subset-superset relationship among $N{:}M$ sparsity levels, we construct the supernet with all $N{:}M$ sparsity choices sharing their weights and applying dynamic masking to extract the sparse matrix, resulting in both reduction of training overhead and improvement of convergence of sparse subnets. With our proposed ELSA methodology, we can yield not only sparsified ViT models with high accuracy but also sparse configurations with effective reduction in FLOPs, which is highly correlated to the speedups yielded by mixed sparsity-supporting accelerators. We anticipate that our work will establish a robust baseline for future research in sparse ViT models with $N$:$M$ sparsity and offer valuable insights for upcoming hardware and software co-design studies.

{
    \small
    \bibliographystyle{ieeenat_fullname}
    \bibliography{main}
}
\newpage
\appendix
\section{Appendix}

\subsection{Details of Our Sparse Supernet Training and Extended Algorithm Pseudocode}
\label{sec:appendix_algo}

Training a supernet is challenging due to the concurrent training of multiple architectures. This complexity necessitates a simplification of the training process to facilitate convergence. Regularization methods (\textit{e.g.,} dropPath, dropout), commonly included in the training process of ViTs, are disabled in our approach. In the dynamic environment of a supernet, where parameters are continuously masked and varied, additional regularization could create training difficulties, obstructing effective learning across the various networks encapsulated within the supernet.

Inspired by prior work in model compression \cite{UVC, DynamicViT}, we incorporate knowledge distillation \cite{DBLP:journals/corr/HintonVD15}, a technique premised on a teacher-student paradigm to further augment and stabilize our supernet training. Specifically, the uncompressed model is the teacher in our implementation. During training, a mini-batch of data $X$ is initially forwarded through this uncompressed model to produce prediction logits. Subsequently, the sparse subnetworks sampled from the supernet are optimized to mimic the uncompressed model's behavior by minimizing the cross-entropy loss relative to the teacher-generated logits. This strategy aims to leverage the guidance from the uncompressed model to faciliate the training of the multiple sparse architectures within our supernet.

We summarize the procedure of our proposed supernet training paradigm and two-step sampling strategy in Algo.~\ref{algo:training_sparse_supernet} and Algo.~\ref{algo:two_ss} (mentioned in Section~\ref{sec:supernet_training}), respectively.

\begin{minipage}[t]{0.49\textwidth}
    \begin{algorithm}[H]
        \caption{Sparse Supernet Training}
        \label{algo:training_sparse_supernet}
        \footnotesize
        \textbf{Input:} 
        \begin{minipage}[t]{0.9\linewidth}
            pretrained weight $\mathcal{W}$, max iteration $T$, \\
            training data $D$, learning rate $\eta$, \\ 
            comput. budget $C_{upper}$, filtering freq. $f_{acf}$
        \end{minipage} \\
        \textbf{Output:} 
        \begin{minipage}[t]{0.8\linewidth}
            trained supernet weights $\mathcal{W}_\mathcal{A}^*$
        \end{minipage}
        \begin{algorithmic}[1]
            \State Init supernet's weight $\mathcal{W}_\mathcal{A}$ with  $\mathcal{W}$
            \State Define computational cost intervals $\mathcal{I}$ by $C_{upper}$
            
            \For{$t$ = $1,\dots,T$}
                \State \textbf{if} ($t$ mod $f_{acf}$ == 0) \textbf{then} Perform \textbf{Algo.~\ref{algo:auto_ch_filter}}
                \State Choose a configuration $\alpha$ by using \textbf{Algo.~\ref{algo:two_ss}}
                \State Derive the sparse weights $\mathcal{W}_{\mathcal{A},t}(\alpha)$ from $\mathcal{W}_{\mathcal{A},t}$
                \Statex \hspace{\algorithmicindent} $\mathcal{W}_{\mathcal{A},t}(\alpha)$ = $\bigl(W^{l}_{sup, t}(s^1),\dots,W^{l}_{sup, t}(s^L)\bigr)$ 
                \State Get a training batch $X$ from $D$
                \State Compute gradient via knowledge distillation 
                \Statex \hspace{\algorithmicindent} $g_t = \nabla \mathcal{L}(X, \mathcal{W}_{\mathcal{A},t}(\alpha), \mathcal{W})$
                \State Update the weights $\mathcal{W}_{\mathcal{A}}$ using gradient descent
                \Statex \hspace{\algorithmicindent}  $\mathcal{W}_{\mathcal{A},t+1} = \mathcal{W}_{\mathcal{A},t}(\alpha) - \eta g_t$
            \EndFor \vspace{-0.4ex}
            \State return trained supernet $\mathcal{W}_{\mathcal{A},T}$
        \vspace{-0.6ex}
        \end{algorithmic}
    \end{algorithm}
\end{minipage}
\hfill
\begin{minipage}[t]{0.49\textwidth}
    \begin{algorithm}[H]
        \caption{Two-Step Sampling}
        \label{algo:two_ss}
        \footnotesize
        \textbf{Input:} 
        \begin{minipage}[t]{0.9\linewidth}
            computational cost intervals $\mathcal{I}$, \\
            current probabilities for sparsity choices $\mathcal{P}_t$
        \end{minipage} \\
        \textbf{Output:} 
        \begin{minipage}[t]{0.8\linewidth}
            a configuration $\alpha$
        \end{minipage}
        \begin{algorithmic}[1]
            \State Uniformly sample an interval $I$ from $\mathcal{I}$ w.r.t
            \Statex \hspace{\algorithmicindent} $I \sim U(\mathcal{I})$ \vspace{-0.5ex}
            \State Choose a configuration \(\alpha\)$=(s^1,\dots,s^L)$ from $I$
            \Statex \hspace{\algorithmicindent} $\alpha \sim \mathcal{P}_t (\{\alpha \mid C_i \leq \mathcal{F}(\alpha) < C_{i+1}\})$
            \Statex \hspace{\algorithmicindent} $/*$ Select $s^l$ from $\mathcal{S}$ based on 
            \Statex \hspace{\algorithmicindent} \hspace{12ex} $\{p_{l,s_1},\dots,p_{l,s_k}\}$ from $\mathcal{P}_t$ $*/$
            \vspace{-0.7ex}
            \State return $\alpha$
            \vspace{-0.3ex}
        \end{algorithmic}
    \end{algorithm}
\end{minipage}

\subsection{Detailed Settings of Sparse Supernet Training}
\label{sec:appendix_setting}
In our experiment, the sparse supernet is trained starting from the pretrained weights of the uncompressed models. The training lasts for 150 epochs. For the supernet training, we mostly keep the original hyper-parameters of each compression target. We employ an AdamW \cite{kingma2017adam} optimizer with an initial learning rate of $5e-5$, a weight decay of 0.005 and a batch size of 1024. We define the computation cost $\mathcal{F}(\cdot)$ as the FLOPs, and user-defined threshold $C_{upper}$ as 50\% FLOPs of the dense models in all supernet training. Also, dropout and dropPath \cite{dropConnect} are both disabled while training the supernet.

\subsection{Other Sparsity Configuration Search Methods in Our Experiments}
\label{sec:appendix_ER_Domino}
As mentioned in Section~\ref{sec:exp_cmp_search}, we compare our proposed method to the following two methods:
\begin{itemize}
   \item \textbf{Erd\H{o}s-R\'enyi (ER)} \cite{ER}: Originally designed to determine layer-wise unstructured sparsity, ER employs a heuristic based on the sum of input and output channels. We modified ER for semi-structured sparsity by initially deciding the unstructured sparsity level and subsequently rounding it to the closest $N$:$M$ sparsity level. For instance, within an $N$:4 search space, if the unstructured sparsity level is 70\%, we would select 1:4 (approximately 75\% zero parameters) as the final semi-structured sparse configuration.

    \item \textbf{DominoSearch} \cite{DOMINO_sun2021dominosearch}: Tailored specifically for CNNs, DominoSearch is an $N$:$M$ sparsity level selection approach. During its operation, each cluster of $M$ consecutive weights in a pruning target (\textit{e.g.,} linear layers) is allocated a threshold, which is determined analytically based on weight magnitude. To delve into layer-wise redundancy, DominoSearch adds an extra regularization penalty, which incrementally pushes the preserved weights towards the predetermined threshold until the desired sparsity is attained.

\end{itemize}

\subsection{Visualization of Search Results}
\label{sec:appendix_visualization}
\begin{figure*}[!htb]
    \centering
    \includegraphics[width=0.9\linewidth]{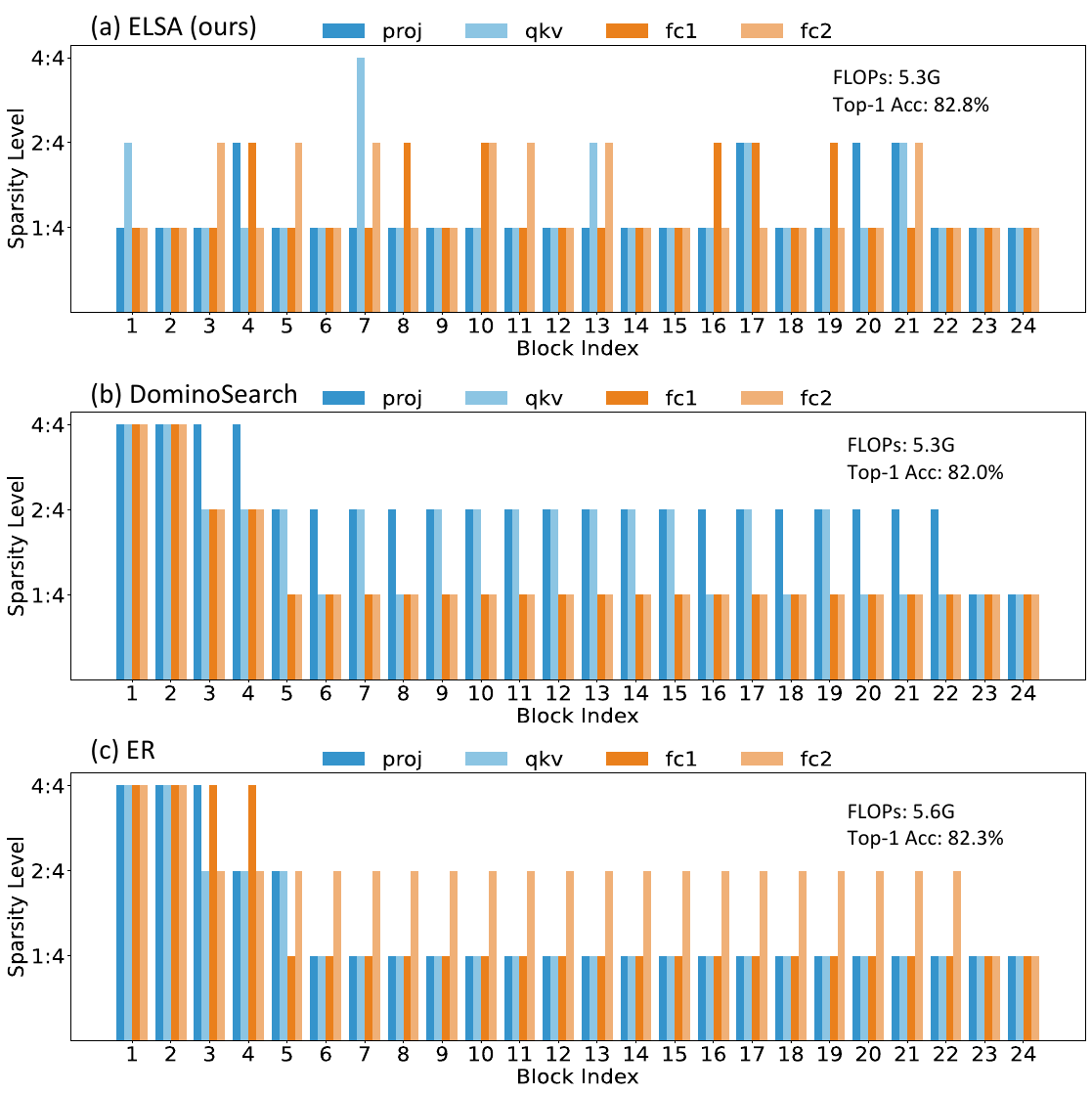}
    \caption{Visualization of sparsity configuration identified by different algorithms for the Swin-B model.}
    \label{Fig:Layer-wise_visualization}
\end{figure*}

In our study, we visually compare the sparsity levels identified by various algorithms for the Swin-Transformer, as depicted in Fig.~\ref{Fig:Layer-wise_visualization}. Observations from Fig.~\ref{Fig:Layer-wise_visualization}(b) and \ref{Fig:Layer-wise_visualization}(c) reveal hierarchical trends in the $N{:}M$ sparsity configuration for compression targets (\textit{i.e.}, weight matrices of linear layers) identified by DominoSearch~\cite{DOMINO_sun2021dominosearch} and ER~\cite{ER}. Specifically, these trends show lower sparsity (applying a denser choice, \textit{e.g.,} 4:4 sparsity) in the layers of the initial blocks, with higher sparsity adopted in the deeper blocks. This pattern aligns with the pyramid-like structure of the Swin-Transformer, where the size of the linear layers increases progressively deeper into the model.

In contrast, the visualization of sparse configuration identified by our proposed ELSA framework, shown in Fig.~\ref{Fig:Layer-wise_visualization}(a), unveils a distinctive pattern. Here, higher sparsity (\textit{e.g.,} 1:4) is applied to the initial blocks, while lower sparsity is selectively employed for the larger-sized linear layers in the model’s middle and deeper sections. This strategic application of sparsity by the ELSA framework leads to a notable 0.8\% improvement in Top-1 accuracy compared to DominoSearch, underscoring the critical importance and benefits of meticulously selecting sparsity levels for each compression target. Furthermore, it showcases the effectiveness of our ELSA framework in navigating these decisions.

\subsection{Results on CNNs}
In this expanded investigation, we have applied the ELSA framework to the ConvNext-S model of convolution neural networks (CNNs) to explore its integration. Our goal is to demonstrate the comprehensive utility and effectiveness of the ELSA framework, showcasing its compatibility not only with vision transformers (ViTs) but also with a broader range of neural network architectures, including CNNs.

\begin{table}[ht]
\def\arraystretch{1.2}
    \centering
    \caption{Experimental results of the proposed ELSA methodology on ConvNext-S. Accuracy denotes the Top-1 accuracy measure on the ImageNet-1K validation set.}
    \resizebox{8cm}{!}{
    \begin{tabular}{c | c c}
        \ChangeRT{1.5pt}
        Model & FLOPs & Accuracy  \\
        \ChangeRT{0.5pt}
        ConvNext-S          & 8.7G                & 82.8\% \\
        ELSA-ConvNext-S-2:4 & 4.3G (1.00$\times$) & 82.3\% \\
        ELSA-ConvNext-S-N:4 & 3.9G (1.11$\times$) & 82.2\% \\
        ELSA-ConvNext-S-N:4 & 3.5G (1.25$\times$) & 82.0\% \\
        ELSA-ConvNext-S-N:4 & 3.1G (1.41$\times$) & 81.6\% \\
        \ChangeRT{1.5pt}
        \end{tabular}
        }
  \label{Tab: ConvNext-S Results}
\end{table}

\begin{figure}[ht]
    \centering
    \includegraphics[width=0.56\linewidth]{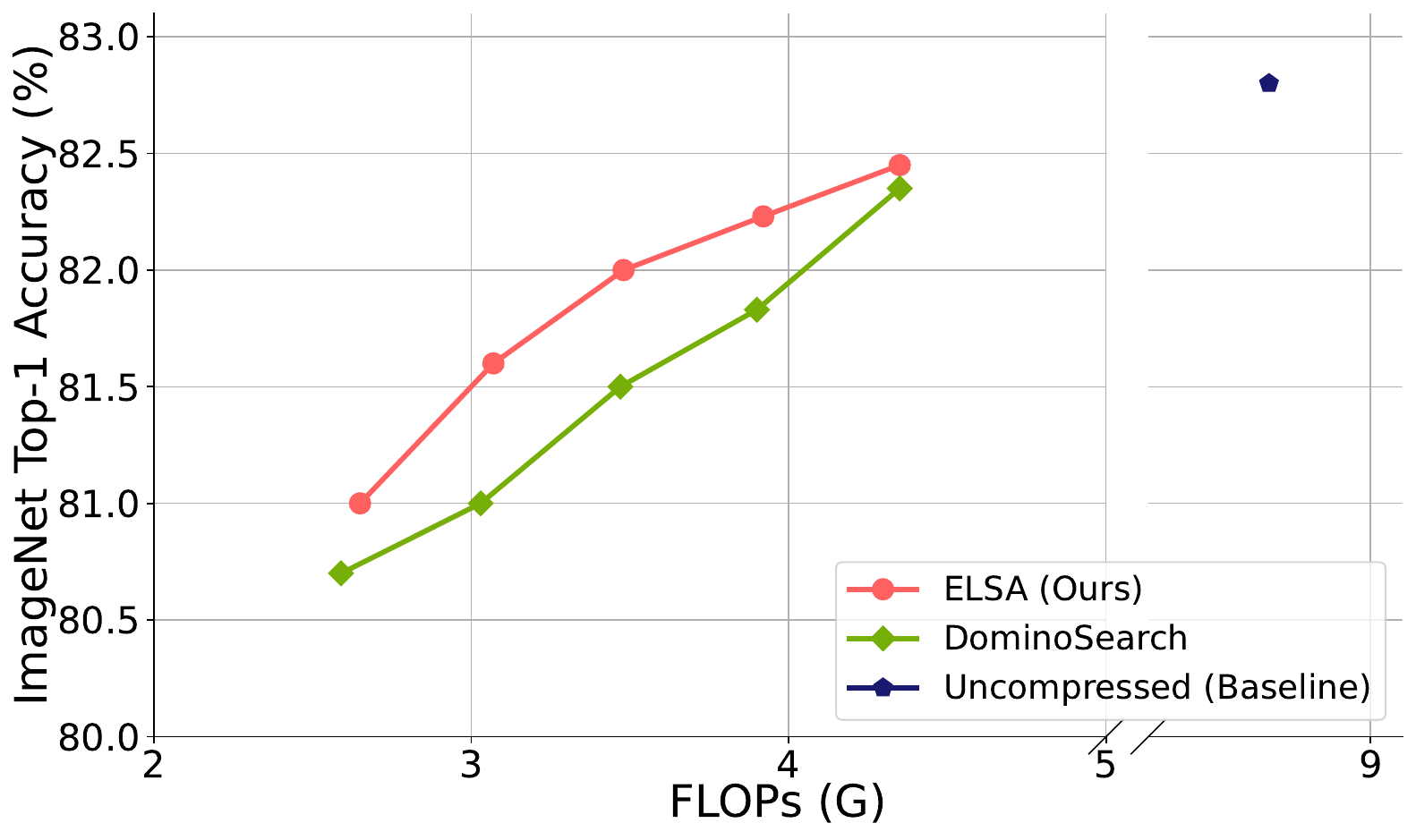}
    \caption{Comparisons of ELSA with DominoSearch on ConvNext-S}
    \label{Fig:ConvNext-S compare}
\end{figure}

To validate this premise, we have conducted experiments on ConvNext-S, a notable advancement in CNN architecture. The experimental results in Table~\ref{Tab: ConvNext-S Results} confirm that the ELSA framework effectively adapts to the ConvNext-S model, significantly enhancing its computational efficiency while maintaining accuracy. This result is particularly significant, highlighting the versatility of the ELSA framework in accommodating various neural network paradigms, despite the operational distinctions between CNNs and ViTs. Furthermore, the $N{:}M$ sparse networks searched by ELSA sit on the Pareto frontier, outperforming DominoSearch, as depicted in Fig.~\ref{Fig:ConvNext-S compare}.

\subsection{Ablation Study and Analysis}

\begin{figure}[ht]
    \centering
    \includegraphics[width=0.5\linewidth]{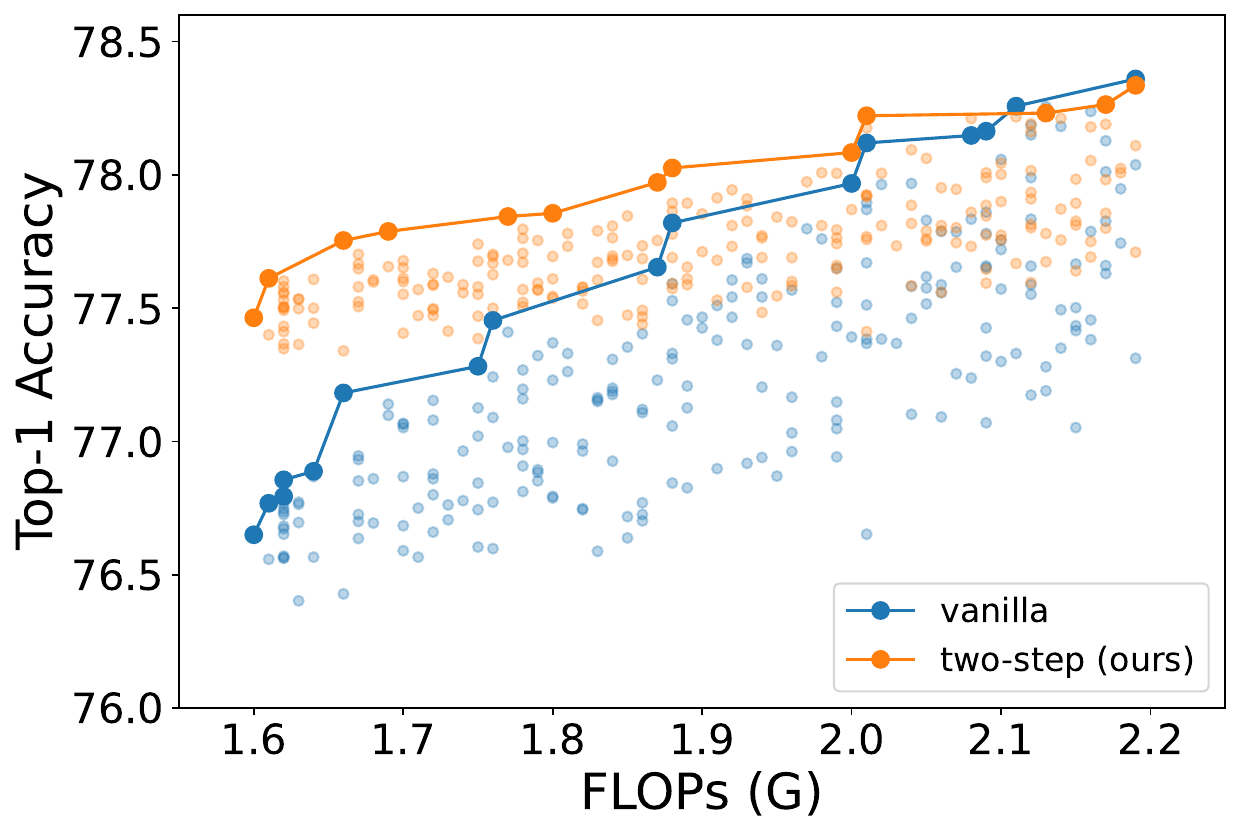}
    \caption{Pareto frontier of subnets randomly sampled from DeiT-S supernet trained with different sampling strategy}
    \label{Fig:Pareto Comparison}
\end{figure}

In this section, we present an ablation study analyzing the impact of our proposed supernet construction and sampling strategy on the resulting supernet quality. We conduct the experiment on the DeiT-S backbone. We visualize the result using the Pareto frontier analysis as shown in Fig.~\ref{Fig:Pareto Comparison}.

\paragraph{Effectiveness of Two-step Sampling Strategy}
In this ablation experiment, we adopt the vanilla sampling strategy~\cite{SPOS}, in which each sparse configuration is sampled uniformly to train a baseline supernet. The significant benefits of our two-step sampling strategy are illustrated in Fig.~\ref{Fig:Pareto Comparison}. Here, we can see that the proposed two-step sampling strategy can yield sparse subnetworks with superior performance at different levels of computation cost. Notably, at around the 1.6G FLOP mark, we can observe a nearly 1\% improvement in Top-1 accuracy.

\paragraph{Effectiveness of Automatic Choice Filtering}
In this ablation experiment, we provide a $N$:8 search space, including five sparsity choices: 1:8, 2:8, 3:8, 4:8, and 8:8, to train two sparse supernets. The baseline supernet only adopts our two-step sampling (TSS) strategy and disables our automatic choice filtering (ACF) strategy. Table~\ref{Tab:acf} shows the effectiveness of our proposed supernet training strategies when a search space with more sparsity choices and even a sparser choice (\textit{i.e.}, the 1:8 sparsity here) is provided. By applying both our two-step sampling and automatic choice filtering, the 1:8 sparsity choice is filtered out from about half of the choice modules during supernet training. Hence, compared to applying two-step sampling only, other sparse choices, such as 2:8, 3:8, and 4:8, have higher probabilities to be chosen. The results indicate that adopting our automatic choice filtering strategy enhances the robustness of supernet training by lowering the probabilities of choices that are too sparse to maintain accuracy.


\begin{table}[ht]
\def\arraystretch{1.2}
    \centering
    \caption{Comparison of the $N$:8 supernets trained with and without our automatic choice filtering strategy being enabled}
    \begin{tabular}{c | r | c c}
    \ChangeRT{1.5pt}
    \multirow{2}{*}{Model} & \multirow{2}{*}{FLOPs} & Trained with both & Trained with only TSS \\
   &   & TSS and ACF & (baseline: disable ACF) \\
    \ChangeRT{0.5pt}
    ELSA-DeiT-B-N:8 &  8.0G & 81.6\% & 80.9\%\\
    ELSA-DeiT-B-N:8 &  7.0G & 81.4\% & 80.5\%\\
    ELSA-DeiT-B-N:8 &  6.0G & 81.1\% & 80.1\%\\
    \ChangeRT{1.5pt}
    \end{tabular}
    
    \label{Tab:acf}
\end{table}

\paragraph{Effectiveness of Supernet in Guiding Searching} 

To demonstrate the benefit of using supernet, we run the evolutionary search and evaluate sparse configurations using lightweight metrics, \textit{i.e.}, the accuracy of the sparse NN, before fine-tuning. As in Table~\ref{Tab: Different Estimator}, the more accurate configuration ranking offered by the supernet can help identify a better sparse configuration with 0.5\% higher accuracy.

\begin{table}[ht]
\def\arraystretch{1.2}
\centering
\caption{Comparison of different estimators (supernet versus dense model)}
  \begin{tabular}{c | r c c}
  \ChangeRT{1.5pt}
   Model & FLOPs & Estimator & Accuracy \\
   \ChangeRT{0.5pt}
   ELSA-DeiT-B-N:4 & 7.0G & Dense model & 81.0\%  \\
   ELSA-DeiT-B-N:4 & 7.0G & Supernet    & 81.5\%  \\
    \ChangeRT{1.5pt}
    \end{tabular}
  \label{Tab: Different Estimator}
\end{table}

\paragraph{Quality of Trained Supernet} 
We aim to construct a high-quality sparse supernet, capable of empowering sparse networks within the design space needed to achieve a performance level close to fine-tuned networks. 
To evaluate our proposed sparse supernet, we employ the training paradigm for $N{:}M$ sparse networks as ASP~\cite{ASP}, refining the performance of sparse subnets through fine-tuning, starting with pretrained models. Each sparse subnet is meticulously fine-tuned for 150 epochs, maintaining consistent settings for knowledge distillation throughout the process. The comparison results are presented in Table~\ref{Tab: From sup vs Finetune}. We note that the sparse subnets derived from our supernet demonstrate only a marginal decline in accuracy, ranging from 0.1\% to 0.2\% across various ViT backbones. These findings emphasize the effectiveness of our supernet training methodology. Through a single training cycle, a broad spectrum of sparse networks can be efficiently generated and inherited from our supernet, substantially reducing the fine-tuning costs.

\begin{table}[ht]
\def\arraystretch{1.2}
\centering
\caption{Comparison of 2:4 sparse subnets with inherited weights and 2:4 finetuned from pretrained weights (150 epochs)}
  \begin{tabular}{c | r | c c}
  \ChangeRT{1.5pt}
   \multirow{2}{*}{Model} & \multirow{2}{*}{FLOPs} & Inherited & Fine-tuned \\
   &   & from ELSA & from pretrained \\
   \ChangeRT{0.5pt}
   DeiT-S & 2.5G & 79.1\% & 79.2\% \\
   DeiT-B & 9.2G & 81.6\% & 81.8\% \\
   Swin-S & 4.6G & 82.8\% & 82.9\% \\
   Swin-B & 8.0G & 83.1\% & 83.3\% \\
    \ChangeRT{1.5pt}
    \end{tabular}
  \label{Tab: From sup vs Finetune}
\end{table}

\subsection{Quantization results}

Our analysis demonstrates that the ELSA framework, combined with quantization methods, maintains the inherent orthogonal characteristics between layer-wise sparsity and quantization compression. According to the results presented in Table~\ref{Tab: Quantization}, it is clear that the reduction in accuracy between compressed ELSA models before and after quantization is similar to that of the dense model before and after.

\begin{table}[ht]
\def\arraystretch{1.2}
\centering
\caption{
Experimental results of the integration of ELSA with quantization technique (PTQ4ViT~\cite{yuan2022ptq4vit}). Accuracy denotes the Top-1 accuracy measure on the ImageNet-1K validation set.}
  \begin{tabular}{ l | c | c | c }
  \ChangeRT{2pt}
    \multirow{2.5}{*}{Model} & \multirow{2.5}{*}{FLOPs} & \multicolumn{2}{c}{Accuracy (\%)}  \\
    \cline{3-4}
    &   &  FP32  &  INT8  \\
    \ChangeRT{1.5pt}
    DeiT-S            & 4.7G & 79.82 & 79.47 (-0.35)  \\
    \ChangeRT{0.25pt}
    ELSA-DeiT-S-2:4   & 2.5G & 79.14 & 78.86 (-0.28)  \\
    \ChangeRT{0.25pt}
    ELSA-DeiT-S-N:4   & 2.0G & 78.34 & 77.77 (-0.57)  \\
    \ChangeRT{1.5pt}
    DeiT-B            & 17.6G & 81.81 & 81.53 (-0.28)  \\
    \ChangeRT{0.25pt}
    ELSA-DeiT-B-2:4   & 9.2G & 81.66 & 81.50 (-0.16)  \\
    \ChangeRT{0.25pt}
    ELSA-DeiT-B-N:4   & 6.0G & 81.37 & 80.96 (-0.41)  \\
    \ChangeRT{1.5pt}
    Swin-S            & 8.7G & 83.17 & 83.15 (-0.02)  \\
    \ChangeRT{0.25pt}
    ELSA-Swin-S-2:4   & 4.6G & 82.81 & 82.68 (-0.13)  \\
    \ChangeRT{0.25pt}
    ELSA-Swin-S-N:4   & 3.5G & 82.53 & 82.44 (-0.09)  \\
    \ChangeRT{1.5pt}
    Swin-B            & 15.4G & 83.45 & 83.34 (-0.11)  \\
    \ChangeRT{0.25pt}
    ELSA-Swin-B-2:4   & 8.0G & 83.10 & 83.14 (+0.04)  \\
    \ChangeRT{0.25pt}
    ELSA-Swin-B-N:4   & 5.3G & 82.80 & 82.71 (-0.09)  \\
    \ChangeRT{2pt}
    \end{tabular}
  \label{Tab: Quantization}
\end{table}

\end{document}